\documentclass[11pt, logo, onecolumn, copyright, colorlinks=true, allcolors=blue]{nvidiatechreport}

\usepackage{graphicx}
\usepackage{subcaption,ragged2e}
\usepackage[round]{natbib}
\usepackage{booktabs}
\usepackage{amsmath}
\usepackage{url}
\title{Accelerating RL Post-Training Rollouts via System-Integrated Speculative Decoding}
\author{Hayate Iso, Tiyasa Mitra, Sudipta Mondal, Rasoul Shafipour, Venmugil Elango, Terry Kong, Yuki Huang, Seonjin Na, Izzy Putterman, Benjamin Chislett, Maor Ashkenazi, Joseph Guman, Gerald Shen, Tugrul Konuk, Ashwath Aithal, Ritika Borkar, Ran Zilberstein, Bita Rouhani}
\date{}

\begin{document}

\begin{abstract}
\large \textbf{Abstract.}
RL post-training of frontier language models is increasingly bottlenecked by autoregressive rollout generation, making rollout acceleration a central systems challenge. Many existing efficiency methods improve throughput by changing the rollout or optimization regime, for example, through off-policy execution, replay, or lower-precision generation. We study speculative decoding as a lossless acceleration primitive for RL rollouts that preserves the target model's output distribution. We implement speculative decoding in NeMo-RL with a vLLM backend, supporting both synchronous and asynchronous pipelines and enabling speculation during RL rollouts. This benefit is realizable across speculation mechanisms, such as pretrained MTP heads, small external draft models or even techniques such as Eagle3, which are traditionally applied after RL phase. This yields a deployment path for state-of-the-art speculative decoding inside RL training. In a reasoning post-training workload at 8B scale under synchronous RL, speculative decoding improves rollout throughput by 1.8$\times$. Using a high-fidelity performance simulator, we project that combining speculative decoding with asynchronous RL yields up to 2.5$\times$ end-to-end training speedup at 235B scale.
\end{abstract}

\maketitle

\section{Introduction}

Reinforcement learning (RL) post-training of frontier language models is increasingly bottlenecked by autoregressive rollout generation. RL is widely used to improve reasoning performance in language models, especially on mathematical and other verifiable tasks \citep{shao2024deepseekmath,DeepSeekAI2025DeepSeekR1IR}, and in these reasoning-oriented workloads overall training time is dominated by rollout generation rather than by gradient computation: across recent large-scale RL systems, rollout generation is routinely the single largest wall-clock component \citep{shen2024nemoaligner,hu2024openrlhf,noukhovitch2024async,piche2025pipelinerl,meta2025llamarl}. The same issue is emerging in agentic RL, where long-horizon tasks require many repeated model calls across tool-use, retrieval, and web-interaction steps and further amplify the cost of every decoded token \citep{jin2025searchr1,qian2025toolrl,wang2025ragen,qi2025webrl}. This makes generation efficiency increasingly important, and speculative decoding offers a way to improve it without changing the verifier-side training semantics.

Several approaches reduce this cost by changing the training dynamics. Asynchronous execution overlaps generation with learning but introduces policy lag \citep{noukhovitch2024async,piche2025pipelinerl,meta2025llamarl}. Off-policy replay and importance-sampling corrections reuse stale trajectories \citep{li2025repo,zheng2025gspo,wang2025aspo,sheng2026espo}. Lower-precision rollouts reduce compute but introduce distribution mismatch \citep{xi2026jetrl,qiu2026fp8rl,li2026qurl}. Selective prompt filtering skips uninformative prompts \citep{zheng2025greso}. These methods are effective, but each changes the sampling or optimization semantics of the original problem.

Speculative decoding offers a different trade-off. A draft model proposes several tokens at once, and the target model verifies them through a rejection procedure that preserves its output distribution \citep{leviathan2023fast,chen2023speculative,cai2024medusa,li2024eagle,mtp2024}. Applied to RL rollouts, this makes generation faster without changing what distribution the trajectories are drawn from, a property that matters because the RL training signal depends on the policy's own samples. In practice, however, the gains depend on how well the draft matches the policy, how many tokens are proposed per step, and what fraction of the RL step is spent on generation.

We study speculative decoding as a rollout acceleration primitive integrated into an open-source RL training framework. We implement it in NeMo RL with a vLLM serving backend, supporting a general drafting path based on EAGLE-3 \citep{li2024eagle,li2025eagle3} that works with any pretrained model, as well as a native path for models with built-in multi-token prediction heads \citep{mtp2024}. The integration handles weight synchronization between the learner and the rollout engine, optional online draft adaptation, and composition with both synchronous and asynchronous RL pipelines.

Our contributions are:
\begin{itemize}
\item We integrate speculative decoding into NeMo RL as a rollout primitive that preserves verifier-exact training semantics, and describe the system requirements (weight synchronization, draft coherence, and stage-level telemetry) needed for a working deployment.
\item We evaluate the integration on 8B-scale reasoning workloads under synchronous RL and study the operational choices (draft initialization, draft length, online adaptation, and interaction with asynchronous execution) that determine realized speedup.
\item We characterize the design space of speculative decoding for RL post-training, quantifying how end-to-end speedup varies with model scale, GPU count, generation share, and policy lag, and identify the operating regimes in which the technique yields meaningful gains at deployment scale.
\end{itemize}

\section{Speculative Decoding for RL Rollouts}
\label{sec:method}

\subsection{Learning speed and lossless acceleration}
\label{sub:learning_speed}

We view RL progress through a simple decomposition \citep{piche2025pipelinerl}:
\[
\text{learning speed} = \text{effectiveness} \times \text{throughput}.
\]
Here, \emph{throughput} is how much rollout and training work the system completes per unit time, and \emph{effectiveness} is how much useful learning signal is extracted from that work. In RL post-training, the training distribution is generated online by the policy itself, so a mechanism that increases throughput by perturbing the sampling distribution may reduce effectiveness. Methods such as asynchronous execution, off-policy replay, and low-precision rollouts each trade effectiveness for throughput in different ways. Speculative decoding targets throughput without changing the sampling distribution: the rejection procedure guarantees that rollouts follow the verifier policy, so effectiveness is preserved by construction.

\subsection{Speedup analysis}
\label{sub:speedup_analysis}

In NeMo RL, a synchronous RL step decomposes as
\[
T_{\text{step}} = T_{\text{data}} + T_{\text{prepare}} + T_{\text{gen}} + T_{\text{logprob}} + T_{\text{train}},
\]
where $T_{\text{prepare}}$ covers weight synchronization and rollout backend preparation, $T_{\text{gen}}$ covers rollout generation, $T_{\text{logprob}}$ is log-probability recomputation under the current policy, and $T_{\text{train}}$ includes advantage computation and policy optimization. Speculative decoding targets only $T_{\text{gen}}$, and within generation, only the autoregressive decode phase (not prefill). The technique is therefore most useful when generation dominates the RL step and generation is decode-heavy.

The step-level speedup is bounded by
\[
S_{\text{step}} \le \frac{1}{R_{\text{gen}} / \alpha + (1 - R_{\text{gen}})},
\]
where $R_{\text{gen}} = T_{\text{gen}} / T_{\text{step}}$ is the generation share and $\alpha$ is the mean acceptance length, defined as the average number of tokens produced per speculation step. This bound follows from Amdahl's law under the assumption that each speculation step costs the same as one autoregressive forward pass. In practice, draft model overhead, prefill time, and batching effects reduce the realized speedup below this bound. The bound nevertheless captures the central trade-off: speculation helps only to the extent that (i) generation dominates step time and (ii) the acceptance length is high enough to outweigh verification cost.

\subsection{System integration}
\label{sub:system}

Integrating speculative decoding into an RL training loop goes beyond adding a draft model to the serving backend. As the policy updates each step, the rollout engine must receive new weights, and the draft model must remain aligned with the current policy. Log-probabilities, KL penalties, and the policy loss must be computed against the verifier (target) policy, not the draft; otherwise, speculative decoding would alter the optimization target. These requirements distinguish RL integration from standalone inference serving, where model weights are fixed.

\begin{figure*}[t]
\centering
  \includegraphics[width=0.98\textwidth]{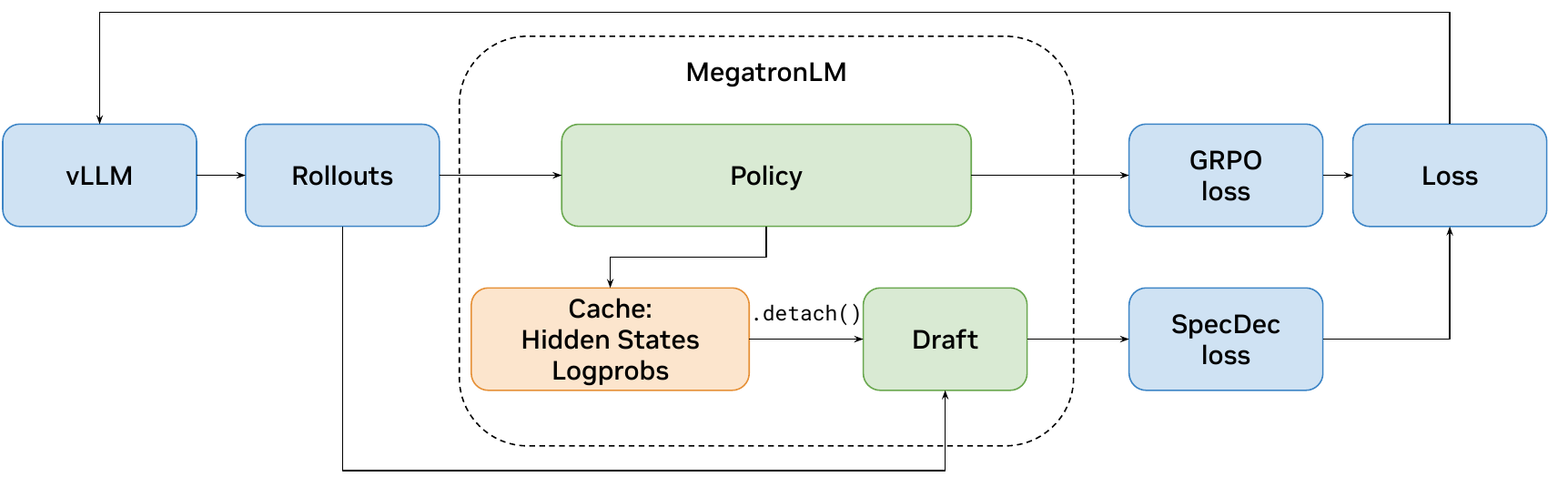}%
\caption{System overview of NeMo RL with speculative decoding. The vLLM backend produces rollout trajectories, and the policy model (MegatronLM) runs the verifier forward pass for the GRPO policy loss. When online adaptation is enabled, hidden states and log-probabilities from that same pass are cached and reused by a separate speculative-decoding loss for the draft head through a gradient-detached pathway (\texttt{.detach()}). The detached boundary ensures that draft training does not alter the policy gradient signal.}
\label{fig:system_overview}
\end{figure*}

Figure~\ref{fig:system_overview} shows the architecture. The vLLM backend generates rollout trajectories using speculative decoding. The policy model (MegatronLM) then runs the forward pass used to compute the GRPO policy loss \citep{shao2024deepseekmath}. When online draft adaptation is enabled, that same forward pass also produces cached hidden states and log-probabilities, which are reused for EAGLE-3 draft supervision rather than recomputed in a separate policy pass. The hidden-state cache is routed through a gradient-detached pathway to the draft head, so that draft training does not interfere with the policy gradient signal.

The system supports two drafting paths. The \emph{general path}, based on EAGLE-3 \citep{li2024eagle,li2025eagle3}, works with any pretrained model without requiring native multi-token-prediction support. A \emph{native path} is also available for models that ship with built-in MTP heads \citep{mtp2024}, where the model's own auxiliary heads serve as the draft mechanism.

Speculative decoding composes with both synchronous and asynchronous RL. In synchronous mode, it directly reduces rollout latency. In asynchronous mode, generation overlaps with other pipeline stages, reducing the effective generation share on the critical path and correspondingly reducing the speedup from speculation. The two mechanisms are complementary: speculation makes each rollout cheaper, while async overlap hides remaining generation cost behind training and log-probability computation.

\section{Experiments}
\label{sec:experiments}

\subsection{Setup}

\emph{\textbf{Policy}.}
We evaluate GRPO-based RL post-training \citep{shao2024deepseekmath} on mathematical reasoning in two settings. \emph{RL-Think} continues training a reasoning-capable model to refine thinking traces, while \emph{RL-Zero} starts directly from the corresponding base model. RL-Think uses Qwen3-8B\footnote{\url{https://huggingface.co/Qwen/Qwen3-8B}}, and RL-Zero uses Qwen3-8B-Base\footnote{\url{https://huggingface.co/Qwen/Qwen3-8B-Base}} \citep{yang2025qwen3}. For training we use DAPO-Math-17K\footnote{\url{https://huggingface.co/datasets/BytedTsinghua-SIA/DAPO-Math-17k}} \citep{yu2025dapo}, and for validation we report accuracy on AIME-2024\footnote{\url{https://huggingface.co/datasets/BytedTsinghua-SIA/AIME-2024}}.

\emph{\textbf{Draft}.}
Our approach leverages the general EAGLE-3 drafting path \citep{li2024eagle,li2025eagle3}, which enables speculative decoding for models that do not have native MTP heads \citep{mtp2024,DeepSeekAI2025DeepSeekR1IR} or specially distilled external draft models. We deliberately focus the empirical study on EAGLE-3 as the harder case: it requires training and maintaining an external drafter, and must be kept aligned with a moving policy throughout RL. The system-level findings below (draft initialization, draft length, online adaptation, and composition with asynchronous execution) carry over directly to the native MTP path \citep{mtp2024}, where the model's built-in auxiliary heads play the role of the draft and the same alignment and configuration trade-offs apply. In the main experiments, we initialize the draft offline using responses generated by the policy on the training prompts, aligning the draft to the actual rollout distribution. Unless otherwise specified, we adopt a draft length of $k=3$ and maintain the EAGLE-3 draft weights fixed during RL. Section~\ref{sub:analysis} further examines the impact of draft initialization, draft length, and online draft adaptation.

\emph{\textbf{Configurations}.}
Experiments run on 8 GB200 NVL72 nodes with 4 GB200 GPUs per node, each with 186\,GB of HBM3E memory, connected by fifth-generation NVLink. We also compare autoregressive decoding against EAGLE-3 and $n$-gram drafting under otherwise matched RL configurations.

\subsection{Main results}
\label{sub:main_results}

We begin by examining where time goes in a typical RL step, since this determines the ceiling for any rollout-side optimization. Table~\ref{tab:main_step_time_breakdown} shows the per-stage breakdown. Generation is the single largest stage, roughly $65$--$72\%$ of step time in both settings, and is the only stage that speculative decoding directly accelerates. Log-probability recomputation and training together account for another ${\sim}27$--$33\%$ and are unchanged regardless of how tokens are generated.

\begin{table}[ht]
  \centering
  \small
  \begin{tabular}{lrrrr}
  \toprule
  & \multicolumn{2}{c}{\textbf{RL-Think}} & \multicolumn{2}{c}{\textbf{RL-Zero}} \\
  \cmidrule(lr){2-3} \cmidrule(lr){4-5}
  Stage & AR (s) & Spec (s) & AR (s) & Spec (s) \\
  \midrule
  Data                 &   0.3 &   0.2 &   0.2 &   0.2 \\
  Prepare              &   2.1 &   1.6 &   1.9 &   2.1 \\
  \textbf{Generation}  & \textbf{133.6} & \textbf{87.0} & \textbf{100.0} & \textbf{56.6} \\
  Log-prob             &  17.9 &  18.1 &  17.8 &  18.1 \\
  Training             &  31.4 &  30.5 &  31.3 &  30.5 \\
  \midrule
  \textbf{Overall} & \textbf{185.3} & \textbf{137.4} ($1.35\times$) & \textbf{151.2} & \textbf{107.5} ($1.41\times$) \\
  \bottomrule
  \end{tabular}
  \caption{Mean step-time breakdown per RL step. Generation accounts for $65$--$72\%$ of step
  time in the autoregressive baseline; speculative decoding reduces it by $1.77\times$ on
  RL-Zero and $1.54\times$ on RL-Think. Log-prob recomputation and training are unchanged,
  capping total step speedup at $1.41\times$ and $1.35\times$ respectively.}
  \label{tab:main_step_time_breakdown}
\end{table}

Given that generation dominates, Table~\ref{tab:draft_config_results} shows how speculative decoding targets this bottleneck. EAGLE-3 reduces generation latency from $100.0$\,s to $56.6$\,s on RL-Zero ($1.8\times$) and from $133.6$\,s to $87.0$\,s on RL-Think ($1.5\times$). These generation-side gains translate to the $1.41\times$ and $1.35\times$ overall step speedups in Table~\ref{tab:main_step_time_breakdown}, bounded by the non-generation stages that speculative decoding cannot touch.

We also compare against $n$-gram drafting, a model-free speculative baseline. Despite achieving non-trivial acceptance lengths ($2.47$ on RL-Zero, $2.05$ on RL-Think), $n$-gram drafting is \emph{slower} than the autoregressive baseline in both settings. This confirms that positive acceptance alone is insufficient; verification overhead can erase the benefit entirely.

\begin{table}[ht]
  \centering
  \small
  \begin{tabular}{l ccc}
  \toprule
  Method & {Accept. len.} & {Gen latency / step (s)} & {Speedup} \\
  \midrule
  \multicolumn{4}{l}{\textbf{RL-Zero}} \\
  Autoregressive & -- & $100.0$ & $1.0\times$ \\
  $n$-gram       & $2.47$ & $140.2$ & $0.7\times$ \\
  EAGLE-3        & $3.32$ & $56.6$  & $1.8\times$ \\\midrule
  \addlinespace
  \multicolumn{4}{l}{\textbf{RL-Think}} \\
  Autoregressive & -- & $133.6$ & $1.0\times$ \\
  $n$-gram       & $2.05$ & $262.9$ & $0.5\times$ \\
  EAGLE-3        & $2.77$ & $87.0$  & $1.5\times$ \\
  \bottomrule
  \end{tabular}
  \caption{Rollout generation comparison. EAGLE-3 substantially reduces generation latency, while $n$-gram drafting is slower than autoregressive decoding despite non-trivial acceptance lengths.}
  \label{tab:draft_config_results}
\end{table}

\begin{figure}[ht]
    \centering
    \begin{subfigure}[t]{\linewidth}
        \centering
        \includegraphics[width=0.9\linewidth]{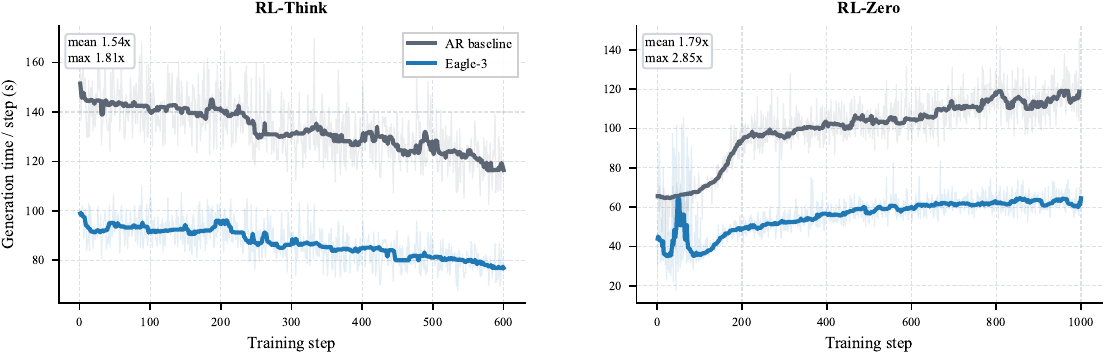}
        \caption{Generation latency per training step. EAGLE-3 achieves mean speedups of $1.54\times$ (max $1.81\times$) on RL-Think and $1.79\times$ (max $2.85\times$) on RL-Zero. The RL-Zero baseline latency rises sharply in the first ${\sim}100$ steps as the policy transitions from short outputs to longer reasoning traces; the EAGLE-3 gap persists throughout.}
        \label{fig:generation_latency}
    \end{subfigure}
    \vspace{0.5em}
    \begin{subfigure}[t]{\linewidth}
        \centering
        \includegraphics[width=0.9\linewidth]{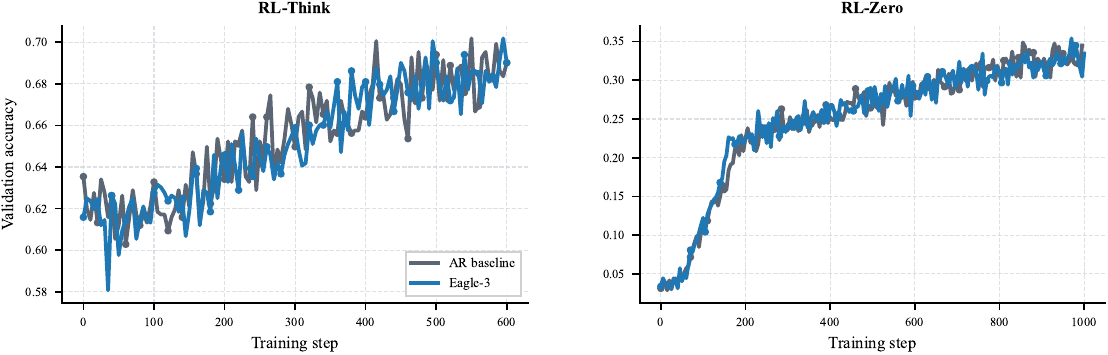}
        \caption{Validation accuracy (AIME-2024) versus training step. The EAGLE-3 and autoregressive curves overlap closely throughout: RL-Think rises from ${\sim}0.60$ to ${\sim}0.70$ and RL-Zero from ${\sim}0.03$ to ${\sim}0.33$, confirming that speculative decoding preserves the optimization trajectory.}
        \label{fig:accuracy_step}
    \end{subfigure}
    \caption{Training dynamics under autoregressive and speculative decoding for RL-Think (left) and RL-Zero (right). (a) Generation latency per step; (b) validation accuracy on AIME-2024. Speculative decoding reduces rollout latency throughout training while producing indistinguishable accuracy curves.}
    \label{fig:training_curves}
\end{figure}

Figure~\ref{fig:training_curves} confirms these findings over the full training run. Generation latency under EAGLE-3 remains consistently below the autoregressive baseline, with mean speedups of $1.54\times$ on RL-Think and $1.79\times$ on RL-Zero (Figure~\ref{fig:generation_latency}). Notably, the RL-Zero baseline latency rises sharply during the first ${\sim}100$ steps as the policy transitions from short outputs to longer reasoning traces; the EAGLE-3 speedup tracks this shift without degradation. Validation accuracy evolves indistinguishably under both decoding modes (Figure~\ref{fig:accuracy_step}), confirming that speculative decoding delivers a faster rollout engine without altering optimization behavior.

\subsection{Analysis}
\label{sub:analysis}

The main results show that speculative decoding speeds up rollout generation without changing the learning trajectory. How much speedup is realized, however, depends on three practical decisions: how the draft model is initialized, how many tokens it proposes, and whether it is updated during training. We study each in turn, then examine the interaction with asynchronous execution as a separate axis.

\paragraph{Draft initialization.}
Table~\ref{tab:init_ablation} shows that draft initialization strongly affects realized speedup. We train an EAGLE-3 draft on UltraChat and Magpie \citep{ding2023ultrachat,xu2025magpie} to obtain a general chat-domain draft, and compare it against a draft trained on the same prompts with DAPO post-training data \citep{yu2025dapo}. At matched draft length $k=3$, the DAPO initialization consistently outperforms the chat draft: speedup rises from $1.51\times$ to $1.77\times$ on RL-Zero and from $1.19\times$ to $1.53\times$ on RL-Think. Initialization quality is not only about generic drafting ability, but about alignment with the rollout distribution encountered during RL.

\begin{table}[ht]
  \centering
  \small
  \begin{tabular}{l cccc}
  \toprule
  \multicolumn{1}{c}{Initialization} & \multicolumn{2}{c}{RL-Zero} & \multicolumn{2}{c}{RL-Think} \\
  & Acceptance & Speedup & Acceptance & Speedup \\\midrule
  UltraChat & $2.88$ & $1.51\times$ & $2.40$ & $1.19\times$ \\
  DAPO & $3.32$ & $1.77\times$ & $2.77$ & $1.53\times$ \\
  \bottomrule
  \end{tabular}
  \caption{Effect of draft initialization at fixed draft length $k=3$ under offline drafting. In-domain initialization improves both acceptance length and realized speedup.}
  \label{tab:init_ablation}
\end{table}

\paragraph{Draft length.}
Table~\ref{tab:length_ablation} studies draft lengths $k \in \{3,5,7\}$. The smallest tested draft, $k=3$, consistently gives the best end-to-end speedup. On RL-Zero, increasing the draft length raises acceptance from $3.32$ to $4.35$ and $5.06$, but speedup falls from $1.77\times$ to $1.44\times$ and $1.21\times$. On RL-Think, speedup drops from $1.53\times$ at $k=3$ to $0.84\times$ at $k=5$ and $0.71\times$ at $k=7$, making the longer drafts slower than autoregressive decoding. Larger drafts increase speculative work enough to erase the benefit of higher acceptance, especially in harder regimes. This finding aligns with the speedup analysis in Section~\ref{sub:speedup_analysis}: the draft-length sweet spot is narrower than the acceptance-length curve would suggest.

\begin{table}[ht]
\centering
\small
\begin{tabular}{c cccc}
\toprule
\multicolumn{1}{c}{Draft length $k$} & \multicolumn{2}{c}{RL-Zero} & \multicolumn{2}{c}{RL-Think} \\
 & Acceptance & Speedup & Acceptance & Speedup \\
\midrule
3 & $3.32$ & $1.77\times$ & $2.77$ & $1.53\times$ \\
5 & $4.35$ & $1.44\times$ & $3.23$ & $0.84\times$ \\
7 & $5.06$ & $1.21\times$ & $3.48$ & $0.71\times$ \\
\bottomrule
\end{tabular}
\caption{Effect of draft length. Acceptance length increases with $k$, but realized speedup declines. For RL-Think, $k \geq 5$ is slower than autoregressive decoding.}
\label{tab:length_ablation}
\end{table}

\paragraph{Online draft adaptation.}
We study online draft adaptation by reusing rollouts generated by the current policy as supervision for the EAGLE-3 draft. Concretely, when online adaptation is enabled, the draft reuses hidden-state and verifier-log-probability caches from the same MegatronLM forward pass that computes the GRPO loss, avoiding an additional policy recomputation. Table~\ref{tab:online_ablation} shows that online updating provides limited additional gains when the draft is already well initialized: DAPO-initialized EAGLE-3 performs nearly identically offline ($1.77\times$) and online ($1.78\times$) on RL-Zero. The larger benefit appears for weaker initialization, where online adaptation improves speedup from $1.51\times$ to $1.63\times$ on RL-Zero and from $1.19\times$ to $1.26\times$ on RL-Think. Online draft maintenance thus serves as insurance against distribution mismatch rather than a general improvement strategy.

\begin{table}[ht]
\centering
\small
\begin{tabular}{l cccc}
\toprule
\multicolumn{1}{c}{} & \multicolumn{2}{c}{RL-Zero} & \multicolumn{2}{c}{RL-Think} \\
Variant & Acceptance & Speedup & Acceptance & Speedup \\
\midrule
UltraChat, offline & $2.88$ & $1.51\times$ & $2.40$ & $1.19\times$ \\
UltraChat, online  & $3.04$ & $1.63\times$ & $2.55$ & $1.26\times$ \\
DAPO, offline      & $3.32$ & $1.77\times$ & $2.77$ & $1.53\times$ \\
DAPO, online       & $3.29$ & $1.78\times$ & $2.74$ & $1.52\times$ \\
\bottomrule
\end{tabular}
\caption{Effect of online draft adaptation at fixed $k=3$. Online updates help the weaker UltraChat initialization, but provide little additional benefit when the draft is already well aligned.}
\label{tab:online_ablation}
\end{table}

\paragraph{Interaction with asynchronous execution.}
The main experiments use synchronous RL to isolate rollout-engine effects. We next revisit the question under asynchronous execution by evaluating RL-Think at policy lag~1 in a 16-node non-colocated configuration, with 12 nodes dedicated to generation and 4 nodes to training. In this regime, much of rollout generation is already hidden behind log-probability recomputation and policy updates, so the relevant quantity is not the full rollout latency but the training-side idle time caused by generation, namely the portion that remains exposed on the critical path. Speculative decoding reduces this exposed generation time from $10.4$\,s to $0.6$\,s per step and lowers effective step time from $75.0$\,s to $60.5$\,s ($1.24\times$). The learning trajectories remain similar under both decoding modes, with speculative decoding preserving the optimization behavior. The gain is smaller than in synchronous RL because asynchronous overlap already hides much of the rollout cost, but the result confirms that speculative decoding and asynchronous execution remain complementary.

\section{Deployment Scale Projections}
\label{sec:projections}

The experiments in Section~\ref{sec:experiments} demonstrate $1.5$--$1.8\times$ generation speedup and up to $1.4\times$ overall step speedup at 8B scale under synchronous RL on 32 GPUs. A natural question is how these gains extrapolate: as model size, GPU count, and policy lag change, does speculative decoding remain effective? Furthermore, while our experimental results capture the overheads of the current software implementation, practitioners may want a software-agnostic view of the maximum speedup opportunity. To bridge this gap, we employ a high-fidelity performance simulator to project the benefits of speculative decoding across a broader deployment design space. The speedups reported in this section should be interpreted as opportunity envelopes, with emphasis on trends rather than absolute values.

\subsection{Simulation methodology}

A faithful simulation framework for RL rollouts must satisfy three requirements: (i) accurately represent device-level and system-level performance characteristics of the serving infrastructure, (ii) support a broad spectrum of model sharding strategies, and (iii) capture the long-tailed response length distributions characteristic of RL rollout workloads.

We utilize a proprietary GPU performance simulator that incorporates detailed models of GPU compute units, memory hierarchies, and interconnects. It leverages a kernel-aware analytical framework to evaluate state-of-the-art model partitioning strategies, including operator overlap and power-aware optimizations. A dynamic traffic generator estimates rollout batch sizes at each step based on a given response length distribution, enabling assessment of speculative decoding's impact on long-tailed rollout latencies across both synchronous and asynchronous configurations.

We study models from the Qwen3 family using a rollout batch size of 4096, scaling to deployments of up to 2048 GB200 GPUs at FP8 precision.

\subsection{Draft length and acceptance length sensitivity}
\begin{figure}[t]
    \centering
    \begin{subfigure}[b]{0.48\linewidth}
        \includegraphics[width=1.0\linewidth]{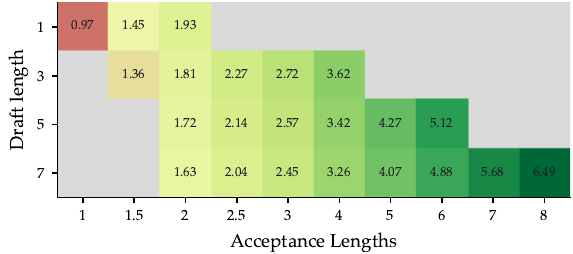}
        \caption{Rollout generation speedup}
    \end{subfigure}
    \begin{subfigure}[b]{0.48\linewidth}
        \includegraphics[width=1.0\linewidth]{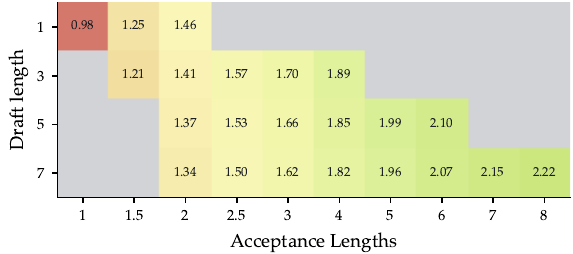}
        \caption{End-to-end RL step speedup}
    \end{subfigure}
    \caption{Simulated rollout (a) and end-to-end (b) speedup for synchronous RL with Qwen3-235B-A22B on 512 GB200 GPUs, as a function of draft length ($y$-axis) and acceptance length ($x$-axis). Gray cells indicate infeasible configurations where acceptance exceeds $k+1$. Non-generation stages bound the peak rollout gain of $6.49\times$ to $2.22\times$ end-to-end.}
    \label{fig:al_dl_heatmaps}
\end{figure}

One key lever for optimizing speedup is the draft length. The optimal choice depends on the achievable acceptance length, which is in turn determined by the dataset and task. Figure~\ref{fig:al_dl_heatmaps} presents a heatmap of expected speedups across a range of draft and acceptance lengths for Qwen3-235B-A22B in a synchronous RL deployment on 512 GPUs.

Longer draft lengths are beneficial only when workloads achieve high acceptance. For example, if $5$ tokens are accepted on average at $k=7$, the longer draft yields a $4.07\times$ rollout speedup but only a $1.96\times$ end-to-end speedup, since non-generation stages dilute the benefit. If acceptance remains at $3$ regardless of draft length, $k=3$ achieves a $2.72\times$ rollout speedup and a $1.70\times$ end-to-end speedup, a comparable end-to-end operating point with far lower speculative overhead. This mirrors the experimental finding in Table~\ref{tab:length_ablation}: higher acceptance length does not automatically justify longer drafts, and the gap between rollout and end-to-end gains grows with draft length.

\subsection{Sensitivity to deployment scale and policy lag}

\begin{figure}[t]
    \centering
    \begin{subfigure}[b]{0.48\linewidth}
        \includegraphics[width=1.0\linewidth]{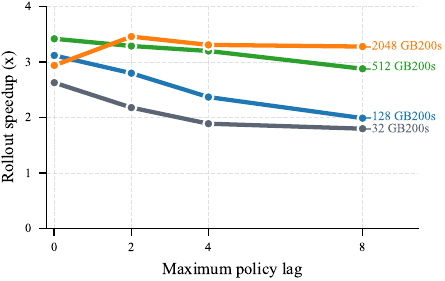}
        \caption{Qwen3-235B-A22B}
    \end{subfigure}
    \begin{subfigure}[b]{0.48\linewidth}
        \includegraphics[width=1.0\linewidth]{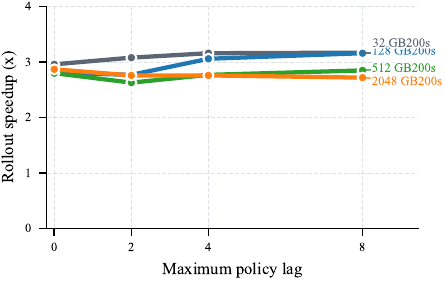}
        \caption{Qwen3-8B}
    \end{subfigure}
    \caption{Simulated rollout speedup across GPU count (32--2048 GB200s) and maximum policy lag (0--8) with draft length 5 and acceptance length 4. (a)~Qwen3-235B-A22B exhibits strong sensitivity to deployment scale: speedup degrades with lag at smaller GPU counts (32, 128) but remains stable at 512--2048 GPUs. (b)~Qwen3-8B is largely insensitive to both scale and lag, with all configurations achieving ${\sim}2.8$--$3.2\times$ rollout speedup.}
    \label{fig:num_gpu_lag_modelsize}
\end{figure}

Asynchronous RL reduces the effective rollout share on the critical path, which limits the opportunity for rollout-side acceleration. However, Figure~\ref{fig:num_gpu_lag_modelsize} shows that substantial gains from speculative decoding remain achievable under realistic async settings. The effect of policy lag is scale-dependent: for the 235B model, smaller deployments (32, 128 GPUs) show pronounced speedup degradation as lag increases, while larger deployments (512, 2048 GPUs) maintain relatively stable rollout speedup even at lag~8. For the 8B model, all configurations are largely insensitive to lag. In all cases, speculative decoding provides meaningful improvement at typical operating lag values.

For sufficiently large models, larger deployments generally derive more benefit from speculation (Figure~\ref{fig:num_gpu_lag_modelsize}a). With a fixed rollout batch size, more GPUs result in a smaller local batch per model instance, exacerbating long-tail under-utilization and creating more room for speculative acceleration. However, this relationship is not monotonic: at zero lag, the 2048-GPU deployment (${\sim}3.0\times$) underperforms 512 GPUs (${\sim}3.4\times$), likely because stretching the batch too thin forces suboptimal sharding. With modest async overlap (lag~2), the 2048-GPU configuration recovers and reaches the highest rollout speedup (${\sim}3.5\times$), as the additional pipeline concurrency compensates for the sharding inefficiency.

The 235B model (Figure~\ref{fig:num_gpu_lag_modelsize}a) shows markedly higher sensitivity to deployment scale and policy lag than the 8B model (Figure~\ref{fig:num_gpu_lag_modelsize}b), where all configurations cluster within a narrow $2.8$--$3.2\times$ band. Larger models occupy more GPUs per instance (e.g., 64 GPUs for 235B versus 8 for 8B in a 2048-GPU deployment) without compromising utilization, while smaller models partition the batch across more instances, fully exposing long-tail effects even at smaller scales. Practitioners training frontier-scale models should therefore pay attention to deployment configuration to maximize the benefit of speculative decoding. At the most favorable simulated operating point (Qwen3-235B-A22B, 2048 GPUs, lag~2), rollout speedup reaches ${\sim}3.5\times$; combined with the high generation share characteristic of frontier-scale models, this translates to a projected ${\sim}2.5\times$ end-to-end training speedup.

\section{Related Work}
\label{sec:related_work}

\paragraph{RL post-training systems.}
Open RL post-training stacks now include NeMo-Aligner, OpenRLHF, veRL, and slime, which provide scalable orchestration and rollout-serving integration \citep{shen2024nemoaligner,hu2024openrlhf,sheng2024hybridflow,verl2025github,slime2025github}. Our work is narrower: rather than a new framework, we study speculative decoding as a deployable rollout primitive inside NeMo RL.

\paragraph{Rollout efficiency.}
Recent systems improve throughput through asynchronous generation, pipelined training, disaggregated engines, replay, importance-sampling corrections, and low-precision rollouts \citep{noukhovitch2024async,piche2025pipelinerl,meta2025llamarl,li2025repo,zheng2025gspo,wang2025aspo,sheng2026espo,xi2026jetrl,qiu2026fp8rl,li2026qurl,zheng2025greso}. Frontier model reports show that RL post-training typically combines several of these levers \citep{deepseek2024v3,yang2025qwen3,minimax2025m1,minimax2025m2,glm2025arc,deepseek2025v32,moonshot2026kimik25,glm2026_5,nvidia2026nemotron3superopen}. Speculative decoding is complementary: it accelerates rollout generation while preserving the target model's sampling distribution.

\paragraph{Speculative decoding for RL.}
Speculative decoding was introduced as a lossless inference accelerator \citep{leviathan2023fast,chen2023speculative} and extended through tree verification, Medusa-style heads, and EAGLE drafting \citep{zhang2024draftverify,cai2024medusa,miao2024specinfer,li2024eagle,li2025eagle3,mtp2024}. Two recent papers apply it to RL specifically. FastGRPO \citep{zhang2025fastgrpo} focuses on concurrency-aware scheduling and online draft learning under high-concurrency group sampling. ReSpec \citep{chen2025respec} studies adaptive draft configurations and reward-weighted drafter adaptation. Our paper differs in focus. We study end-to-end systems integration, including verifier-exact rollout acceleration inside a production-grade RL stack, coordinated weight synchronization, and analysis of how speculation composes with synchronous and asynchronous execution across scales.

\section{Conclusion}
\label{sec:conclusion}

We integrated speculative decoding into NeMo RL as a rollout acceleration primitive that preserves verifier-exact training semantics. On 8B-scale reasoning workloads under synchronous RL, EAGLE-3 speculative decoding reduces rollout generation latency by $1.5$--$1.8\times$ and overall RL step time by up to $1.41\times$, with no change in validation accuracy. Simulator projections show that these gains grow at deployment scale: at favorable operating points for a 235B model, rollout speedups exceed $3\times$ and projected end-to-end training speedup reaches approximately $2.5\times$. Draft initialization quality, draft length, and generation share are the primary determinants of realized gain, and speculative decoding composes with asynchronous execution as a complementary mechanism.

\bibliography{references}
\bibliographystyle{plainnat}

\end{document}